\documentclass{article}
\usepackage{arxiv}

\usepackage{algorithm}
\usepackage{algpseudocode}
\usepackage{xcolor}
\usepackage{natbib}
\usepackage{multirow}
\usepackage{hyperref}
\usepackage{booktabs}

\newcommand{\cmark}{\textcolor{green!60!black}{\checkmark}}
\newcommand{\xmark}{\textcolor{red}{\ensuremath{\times}}}

\usepackage{amsmath,amsfonts,amssymb,amsthm,bm}

\newtheorem{theorem}{Theorem}[section]

\newtheorem{lemma}[theorem]{Lemma}
\newtheorem{proposition}{Proposition}

\def\eqref#1{Equation~\ref{#1}}

\def\1{\bm{1}}

\DeclareMathAlphabet{\mathsfit}{\encodingdefault}{\sfdefault}{m}{sl}
\SetMathAlphabet{\mathsfit}{bold}{\encodingdefault}{\sfdefault}{bx}{n}

\usepackage{amsthm,amsmath,amscd}
\usepackage{amsfonts}
\usepackage{tikz}
 
\usepackage{xspace}

\theoremstyle{definition}

\theoremstyle{remark}
\newtheorem{rmk}{Remark}

\newtheorem*{proposition*}{Proposition}

\PassOptionsToPackage{normalem}{ulem}

\usepackage{enumitem}
\usepackage{graphicx}

\floatstyle{ruled}
\newfloat{algorithm}{tbp}{loa}
\providecommand{\algorithmname}{Algorithm}
\floatname{algorithm}{\protect\algorithmname}

\usepackage{cleveref}

\title{Optimistic Online Learning for Coding Agents with Aggregate Feedback}

\newif\ifuniqueAffiliation

\ifuniqueAffiliation %
\author{Hao~Luan\thanks{Equal contribution.}\\
	School of Computing\\
	National University of Singapore\\
	Singapore, 117417 \\
	\texttt{haoluan@comp.nus.edu.sg} \\
	\And
	Hao~Qiu$^\ast$\\
	School of Computing\\
	National University of Singapore\\
	Singapore, 117417 \\
	\texttt{hao.qiu@nus.edu.sg} \\
    \And
	Zhi~Zheng \\
	School of Computing\\
	National University of Singapore\\
	Singapore, 117417 \\
	\texttt{zhi.zheng@u.nus.edu} \\
    \And
    See-Kiong~Ng \\
    School of Computing \& Institute of Data Science\\
	National University of Singapore\\
	Singapore, 117417 \& 117602 \\
	\texttt{seekiong@nus.edu.sg} \\
    \And
	Zhiyong~Huang \\
	School of Computing\\
	National University of Singapore\\
	Singapore, 117417 \\
	\texttt{dcshuang@nus.edu.sg}
}
\else
\usepackage{authblk}

\author[1]{%
	Hao~Luan\thanks{Equal contribution.}\thanks{Correspondence to Hao~Luan \texttt{<haoluan@comp.nus.edu.sg>}, Hao~Qiu  \texttt{<hao.qiu@nus.edu.sg>}, and Zhiyong~Huang \texttt{<dcshuang@nus.edu.sg>}.}%
}
\author[1]{%
	Hao~Qiu$^\ast$$^\dagger$%
}
\author[1]{%
	Zhi~Zheng%
}
\author[1,2]{%
	See-Kiong~Ng%
}
\author[1]{%
	Zhiyong~Huang$^\dagger$%
}
\affil[1]{School of Computing, National University of Singapore, Singapore 117417}
\affil[2]{Institute of Data Science, National University of Singapore, Singapore 117602}
\fi

\renewcommand{\shorttitle}{Optimistic Online Learning for Coding Agents with Aggregate Feedback}

\date{}

\begin{document}
\maketitle

\begin{abstract}
Large language model (LLM) coding agents operating on Web development platforms can interact with repositories, issue trackers, and continuous integration services to carry out long-horizon software engineering tasks. 
However, behavioral drift remains a major bottleneck in such multi-turn reasoning and action trajectories. 
To mitigate this, activation steering has emerged as a promising approach for regularizing agent behaviors by intervening on the LLM's transformer residual streams. 
Nonetheless, existing methods typically apply a static steering policy across all tasks, overlooking the fact that a single policy is rarely effective across heterogeneous tasks and varying execution turns. 
This limitation motivates dynamic online selection among candidate steering policies. 
Yet, doing so in long-horizon agentic settings is challenging: 
verifiable task-level feedback is only revealed upon task completion, leaving the contributions of individual steering turns unobserved. 
To address this challenge, we formulate online steering-policy selection as a multi-armed bandit problem with delayed task-level feedback and turn-level predictive hints derived from residual-stream activations. 
Within this formulation, we introduce Online Optimistic Activation Steering, a meta-framework that uses hints to guide within-task steering policy selection and leverages terminal task-level feedback to correct cumulative reward estimates. 
Theoretically, against the best fixed policy in hindsight, we prove regret bounds that tighten with increasing hint accuracy and establish a matching lower bound with respect to prediction error and candidate policy count.
Empirical evaluations on realistic coding benchmarks demonstrate that our framework outperforms both static steering policies and naive online adaptive baselines. 
\end{abstract}


\section{Introduction}
\label{sec:intro}
Large language model (LLM) agents can navigate repositories, edit code, and execute tests to address software engineering tasks~\citep{yang2024sweagent}. 
Web-based software development platforms such as GitHub provide APIs for updating pull requests and retrieving automated check results~\citep{githubPullRequestsAPI,githubCheckRunsAPI}.
These multi-turn workflows motivate methods that maintain useful agent behavior throughout task execution.

Activation steering provides a mechanism for influencing LLM behavior by modifying internal activations at inference time~\citep{turner2023steering,wang2025adaptive}.
Recent work extends this approach to coding agents: TACT \citep{sui2026tact} identifies activation directions associated with overthinking and overacting, and intervenes to mitigate these behaviors during task execution.
Existing adaptive steering methods adjust steering directions or intervention strengths based on input semantics~\citep{wang2025adaptive,wang2025semantic}. 
However, the usefulness of a steering intervention may also depend on the task and its current execution stage. 
For example, further investigation may be helpful when diagnosing an unfamiliar failure, whereas repeated investigation after sufficient evidence has been collected can delay a repair.
This motivates online selection among candidate steering policies, using experience across tasks to inform decisions within each task. 

Online policy selection must exploit imperfect predictive signals while learning from task-level aggregate feedback.
Although tool interactions provide intermediate observations, a task outcome does not directly reveal the contribution of each steering decision.
Useful predictive signals may be available before that outcome is known.
For instance, the connection between hidden activations and agent behavior identified by TACT suggests scoring candidate interventions according to how well their resulting activations align with desired behavior~\citep{sui2026tact}.
Such scores can serve as surrogate predictions of reward, which we call \emph{hints}, but behavioral alignment need not translate into task success.
The central challenge is to estimate policy-specific prediction errors when only the combined reward of multiple steering decisions is observed.

We study this problem through a multi-armed bandit model with turn-level predictive hints and task-level aggregate feedback.
An agent executes a sequence of multi-turn tasks and selects one of $K$ candidate activation steering policies at each turn, after observing a hint vector predicting their rewards.
Only the sum of the rewards along the selected policy sequence is revealed at task completion; individual turn-level rewards remain unobserved.
The objective is to minimize regret against the best fixed candidate policy in hindsight.
This additive feedback model isolates the learning problem created by the mismatch between the timing of steering decisions and the granularity of reward observations.

Prior work provides relevant tools, but these require additional mechanisms for the present feedback model.
Adaptive steering methods can vary their interventions with the current input or activation state~\citep{wang2025adaptive,sui2026tact}; this differs from learning which candidate policy to select using aggregate outcomes across tasks.
Bandit methods have also been used for online LLM selection~\citep{xia2024llm}.
A direct task-level EXP3 baseline~\citep{auer2002nonstochastic} selects one policy for an entire task, making its aggregate reward observable, but does not use turn-level hints to adjust decisions within that task. Standard delayed-bandit methods allow feedback to arrive later but assume that each round's selected-arm reward is eventually revealed separately~\citep{cesabianchi2016delay,zimmert2020delays}. Our learner instead observes only a task-level aggregate.
Optimistic online learning instead exploits predictive information to improve regret guarantees~\citep{rakhlin2013predictable,wei2018adaptive}.
Here, the difficulty is to estimate prediction errors when several policies may be applied within a task and only their combined reward is observed.
We are thus motivated by the following question:

\emph{How can an online learner exploit turn-level predictive hints under task-level aggregate feedback, and what regret dependence on prediction accuracy is achievable?}

\textbf{Contributions.}
(i)
We propose \emph{Online Optimistic Activation Steering}, a meta-framework for selecting among existing activation steering policies using turn-level hints and task-level aggregate feedback.
Combining optimistic updates with Tsallis--log-barrier regularization~\citep{rakhlin2013predictable,zimmert2021tsallis,wei2018adaptive}, our algorithm adjusts policy selection within each task and corrects prediction errors at task completion.
Sampling with exponential noise shared across each task provides unbiased estimates of cumulative prediction errors while preserving the intended turn-level policy distributions.
The method supports variable task lengths with a known upper bound $L$. 

(ii)
We establish regret guarantees with leading term $\mathcal{O}(\sqrt{KV})$, where $K$ is the number of candidate steering policies and $V$ measures the cumulative squared prediction error weighted by task length, formally defined in Section~\ref{sec:hint-method}.
This term improves with hint accuracy, unlike the task-level EXP3 guarantee, which is independent of the hints.
The fixed-coefficient algorithm achieves this dependence with tuning based on a prediction error budget (Theorem~\ref{thm:aggregate-fixed}). We also prove a lower bound that matches the leading dependence on the number of policies and prediction error up to universal constants.
Thus, in this regime, our guarantees are minimax optimal when the leading term dominates the lower-order terms omitted here.

(iii) We evaluate our proposed method in simulations and \emph{coding tasks engaged by realistic LLM agents}. Our results show that our method outperforms both static activation steering methods and naive adaptive steering method on the two tested coding benchmarks. 
Simulation experiments verify the advantage of our method in regret bound compared to naive adaptive baseline, and provide insights into our method's sensitivity to hint accuracy, the number of candidate policies, and learning rate.

\section{Related Work}
\label{sec:related}

\paragraph{LLM Agents and Activation Steering.}
LLM agents interleave the model's reasoning and tool use over multi-turn interactions~\citep{yao2023react,yang2024sweagent}. 
Activation steering enables inference-time behavioral control of LLM by modifying hidden representations, from contrastive activation addition~\citep{rimsky2024steering} to context- and state-adaptive interventions~\citep{lee2025programming,zheng2026agentic,cheng2025steering}. 
\citet{davarmanesh2026efficient} further improve steering effectiveness through attention-guided feature extraction and layer selection. 
More recently, TACT~\citep{sui2026tact} applies activation steering to mitigate overthinking and overacting in coding agents. 
Regarding adaptive policy learning, RISER~\citep{ye2026riser} learns to route and compose steering vectors through reinforcement learning with task-level rewards, while feedback-control approaches~\citep{nguyen2026activation,skifstad2026local} dynamically adjust interventions using activation-level errors. 
\citet{wang2025adaptive} also propose adaptive activation steering that dynamically selects steering vectors and adjusts intervention strengths to improve LLM truthfulness, but do not study multi-turn agentic workflows. 
In contrast, our work studies online activation steering policy selection among candidate policies using both turn-level predictive hints and task-level aggregate feedback in long-horizon agentic workflows, and provides regret guarantees that explicitly depend on hint accuracy.

\paragraph{Bandits with Delayed and Aggregated Feedback.}
In bandits with delayed feedback, the learner selects an action at round $t$ and observes its loss only at round $t+d_t$, where $d_t$ is an initially unknown delay~\citep{joulani2013delayed,cesabianchi2016delay,zimmert2020delays}.
Such delays arise in online advertising, where user responses may arrive after an advertisement is displayed.
For adversarial bandits with arbitrary delays, \citet{zimmert2020delays} obtain expected regret $\mathcal{O}(\sqrt{KT}+\sqrt{D\log K})$, where $D\triangleq\sum_{t=1}^{T}d_t$ is the total delay.
In this model, each selected action's loss is eventually revealed separately and associated with its originating round.
Our learner instead observes only the aggregate reward over a known task interval, leaving the contributions of individual steering decisions unobserved. \citet{cesabianchi2022composite} study composite anonymous feedback, where each round reveals a sum of delayed loss components without identifying their originating rounds. Their CoLoWrapper adapts standard bandit algorithms to this feedback. In our setting, feedback is confined to known, non-overlapping task intervals. By committing to one policy throughout each task, its cumulative loss becomes directly observable, allowing EXP3~\citep{auer2002nonstochastic} to run directly at the task level without the wrapper.

We exploit turn-level predictive hints and estimate their cumulative errors from aggregate feedback to obtain regret guarantees that improve with prediction accuracy.

\section{Problem Formulation}
\label{sec:setting}

We first describe the workflow of online activation steering
for coding agents, then formulate the policy-selection problem
as a multi-armed bandit problem with turn-level predictive hints
and task-level aggregate feedback.

\subsection{Online Steering-Policy Selection }

Consider a coding agent addressing a sequence of tasks, such as repository issues requiring fixing bugs or implementing requested changes. 
The finest timescale we adopt for LLM agents in long-horizon tasks is a \emph{turn}, which consists of one round of LLM agent's response including its reasoning and tool call. 
Completing a task often requires multiple turns. for example from 10 to more than 200 in our benchmarks in the experiment section. 
At each turn, a learner selects among candidate activation steering policies, each specifying an intervention in the model's hidden activations. 

\paragraph{Hint observation.}
Before selecting a steering policy, the learner receives predictive scores for the candidate policies.
These hints provide provisional estimates of their rewards before the task outcome is available.
Behavioral scores derived from activations are one possible source of hints.

\paragraph{Policy selection and execution.}
Using the current hints and available history, the learner selects a steering policy to apply during the current turn.
The agent then continues its reasoning or tool interaction.
This selection is repeated across turns within the task.

\paragraph{Task-level feedback.}
At task completion, the learner receives an overall performance signal and uses it to inform subsequent policy selection.
Although tool interactions provide intermediate observations, the final outcome does not directly identify the contribution of each steering decision.
The mathematical model in the next subsection represents reward feedback as an additive task aggregate.

Overall, the goal of this adaptive learner is to select steering policies online to maximize cumulative reward across coding tasks, using turn-level predictive hints and task-level aggregate feedback.

\subsection{Bandit Model with Task-Aggregate Feedback}

We formulate online steering-policy selection as a multi-armed bandit problem with turn-level predictive hints and task-level aggregate feedback. Each arm corresponds to a candidate steering policy in $\Pi_K\triangleq\{\pi_k:k\in[K]\}$, where $K\geq2$. The interaction consists of $B$ consecutive tasks that partition $T$ globally indexed turns. Task $b\in[B]$ spans $\mathcal I_b\triangleq\{s_b,\ldots,e_b\}$ and has length $L_b=e_b-s_b+1$, with $s_1=1$, $s_{b+1}=e_b+1$ for $b<B$, and $e_B=T$; hence $T=\sum_{b=1}^B L_b$. The learner knows a bound $L$ such that $L_b\leq L$ for every task, but knows neither $T$ nor the individual task lengths in advance. Each task endpoint is observed when it occurs.

Let $r_{t,k}\in[0,1]$ denote the reward of policy $\pi_k$ at turn $t$. At each turn $t\in[T]$, the learner first observes a hint vector $m_t\in[0,1]^K$, whose component $m_{t,k}$ predicts $r_{t,k}$. The hints may be inaccurate; our regret guarantees depend on their prediction errors. Using the current hint and available history, it selects $A_t\in[K]$ and applies $\pi_{A_t}$. At the end of task $b$, the learner observes only the aggregate reward $R_b\triangleq\sum_{t\in\mathcal I_b}r_{t,A_t}$; individual turn-level rewards remain unobserved. The reward vectors, hint vectors, and task partition are fixed independently of the learner's randomization.

For each candidate policy $\pi_k$, let $R_{b,k}\triangleq\sum_{t\in\mathcal I_b}r_{t,k}$ denote its cumulative reward on task $b$. The objective is to minimize regret against the best fixed candidate policy in hindsight:
\[
\begin{aligned}
\operatorname{Reg}_T
\triangleq \max_{k\in[K]}\sum_{t=1}^T r_{t,k}
-\mathbb E\!\left[\sum_{t=1}^T r_{t,A_t}\right]=\max_{k\in[K]}\sum_{b=1}^B R_{b,k}
-\mathbb E\!\left[\sum_{b=1}^B R_b\right].
\end{aligned}
\]
The comparator uses the same policy at every turn, chosen in hindsight to maximize cumulative reward.

\paragraph{Notations.}
For a positive integer $n$, let $[n]\triangleq\{1,\ldots,n\}$. 
For a vector $x\in\mathbb{R}^K$, let $x_k$ denote its $k$-th component; for a vector sequence $(x_t)_{t\geq1}$, write $x_t=(x_{t,1},\allowbreak\ldots,x_{t,K})$.
We write $\mathbb{I}\{E\}$ for the indicator of an event $E$.
The probability simplex is
$
    \Delta_K\triangleq\bigl\{p\in\mathbb{R}_{\geq0}^{K}:\sum_{k=1}^{K}p_k=1\bigr\}.
$
Unless stated otherwise, $\mathbb{E}$ denotes expectation over the learner's internal randomization.
Let $\mathcal F_{b-1}$ denote the $\sigma$-algebra generated by the learner's random draws and observations through task $b-1$. We write $\mathbb E_b[\cdot]=\mathbb E[\cdot\mid\mathcal F_{b-1}]$ and $\mathbb P_b(\cdot)=\mathbb P(\cdot\mid\mathcal F_{b-1})$.

\section{Method}

Task-level aggregate feedback does not reveal the contribution of each steering decision. We first present a task-level EXP3~\citep{auer2002nonstochastic} baseline, then introduce Online Optimistic Activation Steering to exploit turn-level hints. We also develop an adaptive variant that requires neither the horizon nor a prediction-error budget. Finally, a lower bound characterizes the worst-case optimality of the leading regret term.

\subsection{Task-Level EXP3 Baseline}
\label{sec:exp3-baseline}
Task-level EXP3~\citep{auer2002nonstochastic} treats each task as one bandit round. By applying a single steering policy throughout the task, it directly observes that policy's cumulative reward.

Within each epoch, the learner maintains $p_b\in\Delta_K$, samples $K_b\sim p_b$ at the start of task $b$, and applies $A_t=K_b$ for all $t\in\mathcal I_b$. At task completion, define $\ell_{b,k}\triangleq L_b-R_{b,k}$ and form the estimate $\widehat\ell_{b,k}\triangleq\allowbreak\mathbb I\{K_b=k\}(L_b-R_b)/p_{b,k}$. The loss-based EXP3 update is $p_{b+1,k}\propto\allowbreak p_{b,k}\exp(-\eta\widehat\ell_{b,k})$, normalized over $k\in[K]$. This update uses only $L_b$ and $R_b$, both available at task completion.

If $\sum_{b=1}^{B}L_b^2$ is known in advance, we use the fixed learning rate $\eta^*\allowbreak=\sqrt{\log K/(K\sum_{b=1}^{B}L_b^2)}$, which minimizes the standard EXP3 regret upper bound~\citep{auer2002nonstochastic}. This is the theoretically tuned baseline used in our experiments. When this quantity is unknown, a doubling trick adapts the learning rate and achieves the same regret rate.\footnote{Play the first task with a uniformly sampled policy and set $H=L_1^2$ upon completion. Each subsequent epoch restarts EXP3 with uniform weights and learning rate $\eta=\sqrt{2\log K/(KH)}$. When the epoch's cumulative squared task lengths exceed $H$ after task $b$, restart with $H=\max\{2H,L_b^2\}$.}

\begin{proposition}
\label{prop:exp3-task-regret}
Both task-level EXP3 with the fixed learning rate $\eta^*$ above and its doubling variant satisfy
\begin{equation}
\label{eq:exp3-task-rate}
\operatorname{Reg}_T=\mathcal{O}\!\left(\sqrt{K\log K\sum_{b=1}^{B}L_b^2}\right).
\end{equation}
The doubling variant requires neither $T$ nor $\sum_{b=1}^{B}L_b^2$ in advance.
\end{proposition}

The guarantee follows from the standard EXP3 analysis~\citep{auer2002nonstochastic} within each epoch, with the first task and each restart-triggering task contributing at most their lengths to regret. Summing over geometrically increasing budgets yields the stated rate without an additional logarithmic factor.

Task-level EXP3 ignores the predictive hints $m_t$ and commits to one policy per task, so its regret guarantee does not improve with hint accuracy. The next method uses hints to adapt policy choices within a task while learning from the same aggregate feedback.

\subsection{Online Optimistic Activation Steering}
\label{sec:hint-method}

Online Optimistic Activation Steering (see Algorithm~\ref{alg:aggregate-hints}) uses hints to select interventions before task feedback becomes available, then corrects its estimates at task completion. This follows the principle of optimistic online learning: exploit reward predictions so that regret depends on prediction errors~\citep{rakhlin2013predictable,wei2018adaptive}. Aggregate feedback creates an additional obstacle: when policies change within a task, their individual contributions cannot be recovered directly. We therefore combine an optimistic Follow-the-Regularized-Leader (FTRL) update with shared randomization and a task-level residual estimator.

The FTRL update in Line~\ref{line:steering-distribution} of Algorithm~\ref{alg:aggregate-hints} combines the score $G_{b-1}$ from completed tasks with accumulated hints and the current hint to compute $p_t$. Accumulated hints supply provisional rewards for the unfinished task, while weighting the current hint by $t-s_b+1$ gives the latest prediction more influence as the task progresses without reward feedback. For $\alpha>0$ and $\beta=6L$, this update uses the regularizer and FTRL map
\[
\begin{aligned}
\psi_{\alpha,\beta}(p)
&=2\alpha\left(\sqrt K-\sum_{k=1}^K\sqrt{p_k}\right)
+\beta\sum_{k=1}^K\log\frac{1}{Kp_k},\\
Q_{\alpha,\beta}(G)
&=\underset{p\in\Delta_K:\,p_k>0}{\arg\max}
\{\langle G,p\rangle-\psi_{\alpha,\beta}(p)\}.
\end{aligned}
\]
The two regularizer components serve complementary purposes. The Tsallis entropy regularizer is motivated by the optimal dependence on the number of arms in Tsallis-INF~\citep{zimmert2021tsallis}. The log-barrier regularizer follows its use for stability in~\citep{wei2018adaptive}. With $\beta=6L$, it limits changes in policy probabilities within a task, keeping the importance weights of our aggregate-feedback estimator under control.

\begin{algorithm}[t]
\caption{Online Optimistic Activation Steering}
\label{alg:aggregate-hints}
\begin{algorithmic}[1]
\Require Task-length bound $L$ and fixed coefficient $\alpha>0$
\State Initialize $\beta=6L$ and $G_0=0$.
\For{each arriving task $b=1,2,\ldots$}
\State For every $k\in[K]$, draw $\xi_{b,k}\sim\operatorname{Exp}(1)$\label{line:steering-noise}
\For{each turn $t$ of task $b$}
\State Update $p_t=Q_{\alpha,\beta}\!\left(G_{b-1}+\sum_{s=s_b}^{t-1}m_s+(t-s_b+1)m_t\right)$\label{line:steering-distribution}
\State Set $A_t=\arg\min_{k\in[K]}\xi_{b,k}/p_{t,k}$ and apply $\pi_{A_t}$\label{line:steering-sampling}
\EndFor
\State Observe $R_b$
\State Set $M_b=\sum_{t\in\mathcal I_b}m_t$ 
\State Set $Z_b=R_b-\sum_{t\in\mathcal I_b}m_{t,A_t}$\label{line:steering-residual}
\State For every $k\in[K]$, set
\State $ \rho_{b,k}=\left(\sum_{\ell=1}^K\max_{t\in\mathcal I_b}\frac{p_{t,\ell}}{p_{t,k}}\right)^{-1}, \widehat D_{b,k}=\frac{\mathbb{I}\{A_t=k\text{ for all }t\in\mathcal I_b\}}{\rho_{b,k}}Z_b.$\label{line:steering-estimator}
\State Update $G_b=G_{b-1}+M_b+\widehat D_b$.\label{line:steering-score-update}
\EndFor
\end{algorithmic}
\end{algorithm}

To learn from task-level aggregate feedback, we exploit tasks in which the same policy is selected throughout, so that the observed reward can be attributed to that policy. We draw exponential noise once at the start of each task and reuse it across turns (Lines~\ref{line:steering-noise} and~\ref{line:steering-sampling} of Algorithm~\ref{alg:aggregate-hints}). Together with the probability stability provided by the log-barrier regularizer, sharing noise makes selecting the same policy throughout a task sufficiently likely. This shared-noise sampling scheme is equivalent to shared Gumbel-max sampling~\citep{oberst2019gumbel,lorberbom2021gumbel}.

Let $D_{b,k}\triangleq\sum_{t\in\mathcal I_b}(r_{t,k}-m_{t,k})$ denote the unobserved cumulative prediction error of policy $k$ on task $b$. Inspired by estimators of \citet{wei2018adaptive}, we use task feedback to estimate and correct hint errors. Line~\ref{line:steering-residual} of Algorithm~\ref{alg:aggregate-hints} computes the task residual $Z_b$, and Line~\ref{line:steering-estimator} applies importance weighting to construct $\widehat D_b$. When policy $k$ is selected throughout task $b$, the observed residual equals its cumulative prediction error, $Z_b=D_{b,k}$. Importance weighting therefore gives $\mathbb E_b[\widehat D_{b,k}]=D_{b,k}$. Consequently, the score update in Line~\ref{line:steering-score-update} adds $M_b+\widehat D_b$, an unbiased estimate of the task reward vector. By estimating prediction errors rather than rewards directly, the estimator's second moment depends on hint accuracy, which leads to our hint-dependent regret bound.

The following theorem bounds the regret of Algorithm~\ref{alg:aggregate-hints} in terms of hint accuracy. We prove it below using the auxiliary lemmas in Appendix~\ref{app:fixed-steering-proof}. The algorithm itself requires no horizon, but the coefficient chosen below depends on the prediction error, which is not observed under aggregate feedback. The next subsection removes the need for this prior information.

\begin{theorem}\label{thm:aggregate-fixed}
Let $V\triangleq\sum_{b=1}^B L_b\allowbreak\sum_{t\in\mathcal I_b}\|r_t-m_t\|_\infty^2$. Choosing $\alpha=\sqrt{24V}+\beta\sqrt K$ ensures that Algorithm~\ref{alg:aggregate-hints} satisfies
\begin{equation}\label{eq:aggregate-tuned-bound}
\operatorname{Reg}_T=\mathcal{O}\!\left(\sqrt{KV}+KL\log\!\left(1+\frac{T}{KL}\right)\right).
\end{equation}
\end{theorem}

\begin{proof}
We use a standard FTRL potential argument~\cite[Section~4.1]{mcmahan2017adaptive}. Define the potential
\[
\Phi_{\alpha,\beta}(G)=\langle G,Q_{\alpha,\beta}(G)\rangle
-\psi_{\alpha,\beta}(Q_{\alpha,\beta}(G)).
\]

Fix $k\in[K]$. To keep the log-barrier finite, use a smoothed comparator~\citep{wei2018adaptive}: set $\gamma=\beta K/(T+\beta K)$ and $u=(1-\gamma)e_k+\gamma\mathbf1/K$, where $e_k$ is the $k$-th standard basis vector and $\mathbf1$ is the all-ones vector. Both regularizer components are nonnegative and vanish at the uniform distribution, so $\Phi_{\alpha,\beta}(0)=0$. Since $u_i\geq\gamma/K$,
\begin{equation}\label{eq:app-comparator-cost}
\psi_{\alpha,\beta}(u)
\leq2\alpha\sqrt K+\beta K\log(1+T/(\beta K)).
\end{equation}
The total reward difference between $e_k$ and $u$ is at most $\gamma T$. Moreover, $\gamma T\leq\beta K\log(1+T/(\beta K))$ by $x/(1+x)\leq\log(1+x)$ with $x=T/(\beta K)$.

We next relate regret to the potential gaps. For task $b$, define
\begin{equation}\label{eq:aggregate-potential-gap}
\delta_b=\Phi_{\alpha,\beta}(G_{b-1}+M_b+\widehat D_b)-\Phi_{\alpha,\beta}(G_{b-1})
-\sum_{t\in\mathcal I_b}\langle p_t,m_t\rangle-Z_b.
\end{equation}
The score updates give $G_B=\sum_b(M_b+\widehat D_b)$. Summing~\eqref{eq:aggregate-potential-gap} and using $\Phi_{\alpha,\beta}(0)=0$ yields
\[
\begin{aligned}
\Phi_{\alpha,\beta}(G_B)
&=\sum_b\delta_b + \sum_b\left(\sum_{t\in\mathcal I_b}\langle p_t,m_t\rangle+Z_b\right).
\end{aligned}
\]
The variational definition gives $\Phi_{\alpha,\beta}(G_B)\geq\langle G_B,u\rangle-\psi_{\alpha,\beta}(u)$, so
\begin{equation}\label{eq:app-potential-regret}
\sum_b\left\langle u,M_b+\widehat D_b\right\rangle -\sum_b\left(\sum_{t\in\mathcal I_b}\langle p_t,m_t\rangle+Z_b\right)\leq\psi_{\alpha,\beta}(u)+\sum_b\delta_b.
\end{equation}
By Lemma~\ref{lem:steering-moments}, the comparator's estimated reward satisfies
\[
\mathbb E_b\!\left[\langle u,M_b+\widehat D_b\rangle\right]
=\langle u,M_b+D_b\rangle=\sum_{t\in\mathcal I_b}\langle u,r_t\rangle.
\]
For the learner, the definition of $Z_b$ and the marginal identity $\mathbb P_b(A_t=k)=p_{t,k}$ in Lemma~\ref{lem:steering-coupling} give
\[
\mathbb E_b[Z_b]
=\mathbb E_b[R_b]-\sum_{t\in\mathcal I_b}\langle p_t,m_t\rangle.
\]
Thus the expected hint contribution cancels, yielding
\[
\mathbb E_b\!\left[\sum_{t\in\mathcal I_b}\langle p_t,m_t\rangle+Z_b\right]
=\mathbb E_b[R_b].
\] We take expectations in~\eqref{eq:app-potential-regret} and apply Lemma~\ref{lem:steering-gap}. Here $V_b\triangleq L_b\sum_{t\in\mathcal I_b}\|r_t-m_t\|_\infty^2$, so $V=\sum_{b=1}^B V_b$. Adding the smoothing cost yields
\begin{equation}\label{eq:app-fixed-explicit}
\begin{aligned}
\operatorname{Reg}_T
\leq2\alpha\sqrt K+2\beta K\log(1+T/(\beta K)) +\frac{48\sqrt K}{\alpha}V.
\end{aligned}
\end{equation}
Because the reward sequence is fixed, a best fixed arm can be chosen independently of the learner's randomization. The same inequality therefore bounds regret against the best fixed arm.

Choosing $\alpha=\sqrt{24V}+\beta\sqrt K$ and $\beta=6L$ gives~\eqref{eq:aggregate-tuned-bound}.
\end{proof}

\begin{rmk}
\label{rmk:delayed-bandits}
The leading regret term in Theorem~\ref{thm:aggregate-fixed} is $\mathcal{O}\bigl((K\sum_{b=1}^B L_b\allowbreak\sum_{t\in\mathcal I_b}\allowbreak\|r_t-m_t\|_\infty^2)^{1/2}\bigr)$.
It is at most $\mathcal{O}(\sqrt{K\sum_{b=1}^B L_b^2})$ and becomes smaller as prediction errors shrink. In activation steering, hints provide a basis for choosing among candidate interventions before a coding task finishes. Under the analyzed model, more accurate reward predictions improve the leading guarantee against the best fixed steering policy, even though the contribution of each intervention is not separately observed.

In standard delayed bandits, the reward of the selected arm at each round is eventually revealed separately, together with its round index. \citet{zimmert2020delays} obtain $\mathcal{O}(\sqrt{KT}+\sqrt{D\log K})$ regret, where $D$ is the total delay. If individual turn-level rewards were revealed at task completion in our setting, then $D=\frac12\sum_{b=1}^B L_b(L_b-1)$. Our learner instead observes only their sum, preventing direct use of their turn-level reward estimators. Our guarantees exploit predictive hints to reduce regret under task-level aggregate feedback.

\end{rmk}

\subsection{Adaptive Optimistic Steering}
\label{sec:adaptive-hint-method}
The coefficient in Theorem~\ref{thm:aggregate-fixed} depends on cumulative prediction errors that aggregate feedback does not reveal. Adaptive Online Optimistic Activation Steering (Algorithm~\ref{alg:aggregate-adaptive}) removes this need for prior information through feedback-driven restarts. It keeps regularization fixed within each epoch and strengthens it when an observable cumulative potential gap exceeds a threshold. This lets the learner adapt its effective learning rate through the Tsallis regularization coefficient, without knowing the horizon or a prediction-error budget. The design is motivated by the mixability-gap budget and restart construction of AdaHedge~\citep{vanerven2011adaptive}. We use a task-level potential gap computed from hints and
aggregate feedback to adapt the Tsallis regularization coefficient.

Let $H_b$, $G_b$, and $S_b$ denote the threshold, score, and cumulative signed gap after task $b$, including any restart. Task $b$ uses $\alpha_b=H_{b-1}/(4\sqrt K)$, with $\beta=6L$ fixed. Both coefficients remain constant within each epoch. At task completion, compute the potential gap in~\eqref{eq:aggregate-potential-gap} with $\alpha=\alpha_b$.
The potential gap compares the increase in regularized potential with an unbiased estimate of the reward earned during the task. It is observable at task completion, and its conditional mean is bounded in terms of prediction error, allowing the learner to decide when to increase the Tsallis regularization coefficient $\alpha$ and restart. If its cumulative sum exceeds the threshold, the algorithm doubles the threshold and restarts with stronger Tsallis entropy regularizer (Lines~\ref{line:adaptive-threshold}--\ref{line:adaptive-restart} of Algorithm~\ref{alg:aggregate-adaptive}). Keeping $\beta$ fixed preserves the stability needed for the aggregate-feedback estimator. 
Each task uses Algorithm~\ref{alg:aggregate-hints}'s sampling and estimator with the supplied score and coefficients, drawing fresh noise and sharing it across its turns.

\begin{algorithm}[t]
\caption{Adaptive Online Optimistic Activation Steering}
\label{alg:aggregate-adaptive}
\begin{algorithmic}[1]
\Require Task-length bound $L$ and Algorithm~\ref{alg:aggregate-hints} as a subroutine
\State Initialize $\beta=6L$, $H_0=4\beta K$, and $G_0=S_0=0$.
\For{each arriving task $b=1,2,\ldots$}
\State Set $\alpha_b=H_{b-1}/(4\sqrt K)$.
\State Execute the task-$b$ steps of Algorithm~\ref{alg:aggregate-hints} with input $(G_{b-1},\alpha_b,\beta)$ to obtain $(M_b,\widehat D_b,Z_b,\{p_t\}_{t\in\mathcal I_b})$, without updating the score.
\State At completion, compute $\delta_b$ using~\eqref{eq:aggregate-potential-gap}.
\If{$S_{b-1}+\delta_b>H_{b-1}$}\label{line:adaptive-threshold}
\State Set $H_b=2H_{b-1}$ and $G_b=S_b=0$.\label{line:adaptive-restart}
\Else
\State Set $H_b=H_{b-1}$, $G_b=G_{b-1}+M_b+\widehat D_b$, and $S_b=S_{b-1}+\delta_b$.
\EndIf
\EndFor
\end{algorithmic}
\end{algorithm}

The following theorem shows that Algorithm~\ref{alg:aggregate-adaptive} preserves the hint-dependent leading regret term without knowing $T$ or a pre\-diction-error budget. Its proof is given in Appendix~\ref{app:adaptive-steering-proof}.

\begin{theorem}\label{thm:aggregate-adaptive}
Under the assumptions of Theorem~\ref{thm:aggregate-fixed}, with $V$ defined there, Algorithm~\ref{alg:aggregate-adaptive} requires neither $T$ nor a prediction-error budget and satisfies
\begin{equation}\label{eq:aggregate-adaptive-bound}
\operatorname{Reg}_T=\mathcal{O}\!\left(\sqrt{KV}+KL\log\!\left(1+\frac{T}{KL}\right)\left[1+\log\!\left(1+\frac{T}{KL}\right)\right]\right).
\end{equation}
\end{theorem}
Compared with the optimally tuned version in Theorem~\ref{thm:aggregate-fixed}, the adaptive algorithm requires no prior knowledge of the prediction error and retains the $\mathcal{O}(\sqrt{KV})$ leading term, at the cost of an additional logarithmic factor in the additive term. This extra factor arises from summing the regularization costs over the epochs created by the doubling trick.

\subsection{Lower Bound and the Scope of Optimality}
\label{sec:steering-lower-bound}
The following theorem establishes a worst-case regret lower bound. Its proof is given in Appendix~\ref{app:steering-lower-bound}.

\begin{theorem}
\label{thm:steering-lower-bound}
Let $B\geq K\geq2$ and $L\geq1$ be integers, and let $T=BL$. For any prediction-error budget $0<\overline V\leq BL^2/4$ and any randomized learner receiving turn-level hints and only task-level aggregate rewards, there exists a fixed instance with $B$ tasks of length $L$ and reward and hint sequences satisfying $V\leq\overline V$, where $V$ is defined in Theorem~\ref{thm:aggregate-fixed}, such that
\begin{equation}\label{eq:steering-budget-lower}
\operatorname{Reg}_T=\Omega\!\left(\sqrt{K\overline V}\right).
\end{equation}
The result holds even if the learner knows $B$, $L$, and $\overline V$ in advance.
\end{theorem}

Together with Theorems~\ref{thm:aggregate-fixed} and~\ref{thm:aggregate-adaptive}, this lower bound shows that the leading $\sqrt{K\overline V}$ dependence is optimal in the worst case. The full regret guarantees are minimax optimal when this term dominates their respective additive terms. This lower bound shows that the dependence on prediction error and the number of candidate policies is unavoidable, even when the learner can switch policies within each task.

\section{Experiments}
\label{sec:exps}

Through empirical evaluations in simulations and real LLM agents on coding benchmarks, we answer the following research questions: 
\begin{enumerate}[leftmargin=*,itemsep=2pt,topsep=-2pt]
    \item Is a fixed steering policy optimal for heterogeneous long-horizon coding tasks? 
    \item Is our method more effective for LLM agents in real coding tasks by combining both activation hints and task-level feedback for adaptive activation steering, compared to baselines? 
    \item How does our method perform asymptotically when given enough interactions compared to the most naive adaptive baseline? How sensitive is it against hint accuracy, number of candidate policies, and learning rate? 
\end{enumerate}

\subsection{Is a Static Steering Policy Enough?}
\label{sec:lambda-nonstationarity}

We begin with a case study to illustrate that in practice, one single steering policy \emph{cannot} perform optimally on all tasks. 
Thus, this motivates an adaptive meta-framework that can adaptively select different policies at least for different tasks. 

\paragraph{Setup.}
We apply static TACT-Gate steering \citep{sui2026tact} on 4 selected tasks from SWE-Bench Verified (SWE-V) \cite{jimenez2024swebench} with different levels of difficulties (\texttt{Astropy-12907}, \texttt{-13236}, \texttt{-14365} and \texttt{-14096}).
Based on TACT-Gate, we derive 5 different static policies by varying one of its hyperparameter $\lambda \in \{0.25, 0.5, 1.0, 1.5, 2.0\}$, where $\lambda=1.5$ was reported the best in \citep{sui2026tact}. 
We then run all 5 static policies on the 4 selected tasks, yielding 20 policy-task combinations. 
With all the 20 combinations, we therefore have a hypothetical ``adaptive oracle'' that can choose the best policy for each task \emph{in hindsight}.

\begin{table}[t]
\centering
\caption{Resolving rates of different static steering policies in the 4-task case study (\cmark{} = resolved, \xmark{} = unresolved).}
\label{tab:lambda-grid}
\small
\begin{tabular}{lcccccc}
\toprule
Policy $\backslash$ Task & T1 & T2 & T3 & T4 & Resolved \\
\midrule
$\lambda=0.25$ & \cmark & \xmark & \cmark & \xmark & 2/4 \\
$\lambda=0.5$  & \cmark & \xmark & \cmark & \xmark & 2/4 \\
$\lambda=1.0$  & \cmark & \xmark & \xmark & \xmark & 1/4 \\
$\lambda=1.5$  & \cmark & \xmark & \xmark & \cmark & 2/4 \\
$\lambda=2.0$  & \cmark & \xmark & \cmark & \xmark & 2/4 \\
\midrule
adaptive oracle & \cmark & \xmark & \cmark & \cmark & \textbf{3/4} \\
\bottomrule
\end{tabular}
\end{table}

\paragraph{Results.}
On the 4-task mini benchmark, 4 out of 5 static policies score the same resolve rate. 
T1 is easy enough that all policies can solve it, and T2 the opposite. 
Interestingly, $\lambda=1.5$ trades task T3 for T4 which no other static policies can resolve. 
The intersection of the task-optimal $\lambda$ sets is empty, so no single $\lambda$ is optimal for every task on this grid. 
However, if a non-static steering policy can adaptively choose different static policies on the fly at least for different tasks, as the adaptive oracle does, it is possible to achieve the best resolve rate ($3/4$) in hindsight that no static policy could ever achieve on this grid. 
Taken together, this suggests that a single static steering policy appears unable to be optimal across all tasks in practice, and therefore motivates online adaptive policy selection for better empirical performance.

\subsection{Adaptive Steering on Real Coding Tasks}

\begin{table}[t]
    \centering
    \caption{Task resolve rate comparison across all methods on SWE-V and TB2. Resolve rate is the higher the better.}
    \label{tab:agent_results}
    \begin{tabular}{lcc}
        \toprule
        \textbf{Method} & \textbf{SWE-V (\%)} $\uparrow$ & \textbf{TB2 (\%)} $\uparrow$ \\
        \midrule
        Vanilla Decoding      & 55.0 & 28.1 \\
        Static TACT-Cap       & 62.0 & 32.6 \\
        Static TACT-Gate      & 57.0 & 30.3 \\
        B-EXP3-TACT-Cap (adaptive)      & 62.0 & 31.5 \\
        B-EXP3-TACT-Gate (adaptive)     & 58.0 & 27.0 \\
        \midrule
        Ours-TACT-Cap (adaptive)         & \textbf{65.0}   & \textbf{34.8}\\ 
        Ours-TACT-Gate (adaptive)        & \textbf{63.0}   & \textbf{32.6} \\
        \bottomrule
    \end{tabular}
\end{table}

We instantiate our online adaptive framework on a recent activation steering method TACT \citep{sui2026tact}, and demonstrate that our method benefits from combining activation hints and verifiable task feedback in terms of improving coding agents' performance on realistic coding benchmarks. 

\paragraph{Baselines and metrics.}
We compare our method with both static steering policies and naive adaptive method. 
Static steering baselines include vanilla decoding, where zero activation steering is applied, and TACT with two of its variants, Gate and Cap. 
The adaptive task-level EXP3 (B-EXP3) with doubling trick mentioned in \cref{sec:exp3-baseline} is our adaptive baseline. 
We adopt the resolve rate (portion of resolved tasks among all tested) as the sole metric to evaluate methods' performance. 

\paragraph{TACT-related setups.}
We mostly follow\footnote{
We reproduced TACT to the best from the contents disclosed in the paper, since its official implementation is not yet publicly available until submission of this manuscript. 
} 
the setups disclosed in \citep{sui2026tact}.  
We test all compared methods on two benchmarks: SWE-V \citep{jimenez2024swebench} and Terminal Bench 2.0 (TB2) \citep{merrill2026terminalbench}, containing 500 and 89 tasks, respectively. 
For both benchmarks, we adopt \texttt{mini-swe-agent} \citep{yang2024sweagent} as the coding agent harness and use public Qwen3.5-27B \citep{qwen35blog} as the agent's base LLM. 
We follow \citet{sui2026tact}'s layer selection protocol to steer 10 out of all 64 layers of the LLM model. 
We adopt the optimal hyperparameters reported in \citep{sui2026tact} for both TACT-Cap and -Gate. 

\paragraph{Online adaptation setup.} 
To make TACT adaptive, we discretize the continuous hyperparameters' spaces into grids for each TACT variant so the candidate class of policies is parameterized.  
For TACT-Cap, we discretize its capping boundary in the unit of activations' calibrated projection standard deviation; specifically, we use 8 candidate policies $k\in\{1.0, 1.25, 1.5, 1.75, 2.0, 2.25, 2.5, 3.0\}$ on SWE-V and $k\in\{1.6, 1.7, 1.8, 1.9, 2.0, 2.1, 2.2, 2.3\}$ on TB2, covering the hyperparameter search range reported in \citep{sui2026tact}. 
Likewise for TACT-Gate, we discretize the correction strength of steering: we set $\lambda\in\{0.1, 0.25, 0.5, 0.75, 1.0, 1.25, 1.5, 2.0\}$. 
The max task lengths is set to $L=256$ on SWE-V and $L=128$ on TB2 for all methods. 
For our method, we only evaluate Algorithm~\ref{alg:aggregate-adaptive} in this real agent setting due to limited compute budget. 

\paragraph{Hint setup.}
Following TACT, we first define a probe function that maps the live hidden state of each selected layer $\ell$ to standardized coordinates $z_\ell = \mathrm{diag}(\sigma_\ell)^{-1} D_\ell (h_\ell - \mu_\ell) \in \mathbb{R}^2$, where $h_\ell$ denotes the live residual-stream hidden state, $\mu_\ell$ is the calibration mean of $h_\ell$, $D_\ell$ represents the two calibrated probe directions (axes) at layer $\ell$, and $\sigma_\ell$ is the calibration standard deviations along the two axes. 
Then, we predict the post-intervention coordinates analytically for each candidate steering policy. 
For TACT-Cap, 
$$
\tilde z_\ell = \mathrm{clip}(z_\ell,-k,k), 
$$
and for TACT-Gate, 
$$
\tilde z_\ell = z_\ell + \lambda(\mathrm{clip}(z_\ell,-k,k) - z_\ell). 
$$ 
With the above, we define the hint as the layer-averaged soft box reward of these coordinates: 
$$
m_A = \frac{1}{\sum_\ell 1}\sum_\ell \prod_{j} \sigma(\gamma \, g_j(\tilde z_\ell)) \in [0,1]
$$ 
where $\sigma(\cdot)$ is the logistic sigmoid function, $g_j$ represents the four signed distances to the box edges, and $\gamma>0$ is a sharpness scale which we set as $10.0$ in our experiments. 
Together, the hint approximates how well the activations stay in the calibrated band defined by TACT.

\paragraph{Results.} 
The main results are presented in Table~\ref{tab:agent_results}. 
Static TACT-Cap and -Gate improves over vanilla decoding by $2\%$--$7\%$.  
This suggests that the TACT steering can help regulate the agent's behaviors in a way that facilitates resolving coding tasks on the evaluated benchmarks, which is consistent with the conclusion reported in \citep{sui2026tact}. 
The task-level EXP3, however, does not necessarily show improvements: B-EXP3 improves over static TACT-Gate by only $1\%$ and ties with static TACT-Cap on SWE-V, while yielding worse performance on TB2 when adapting TACT-Gate. 
This appears to suggest that task-level aggregate feedback alone without the extra information revealed by activation hints may even result in jeopardizing the overall performance by committing to suboptimal steering policies during an entire task. 
In contrast, our method receives performance gains on both tested benchmarks over the compared baselines, thanks to combining both turn-level activation hints and task-level aggregate feedback and using them to adaptively select different steering policies at each turn when the agent is in the process of attempting each task.

\subsection{Simulation Study and Ablations}
\label{sec:sim}

\begin{figure}[t]
\centering
\includegraphics[width=\columnwidth]{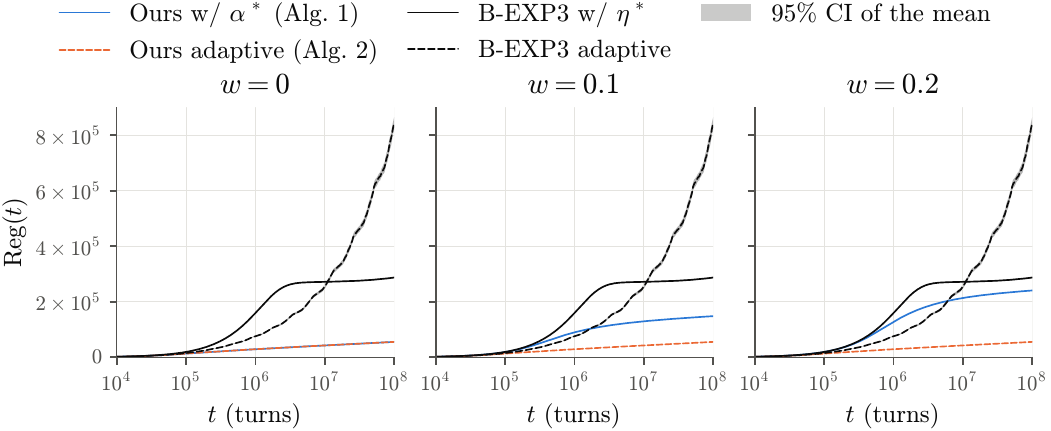}
\caption{Regret curves over turns $t$ with varied hint noise levels $w\in\{0,0.1,0.2\}$.}
\label{fig:e1}
\end{figure}

\begin{figure*}[t]
\centering
\includegraphics[width=0.99\textwidth]{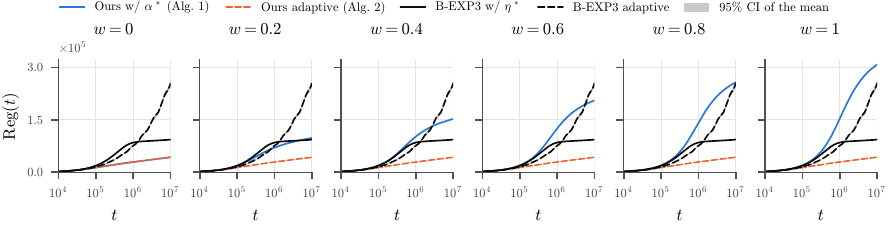}
\caption{Ablation in hint accuracy: Regret vs.\ $t$ for each tested noise level $w$ ($T{=}10^7$). Bands indicate 95\% CIs.}
\label{fig:e2curves}
\end{figure*}

\begin{figure*}[t]
\centering
\includegraphics[width=0.98\textwidth]{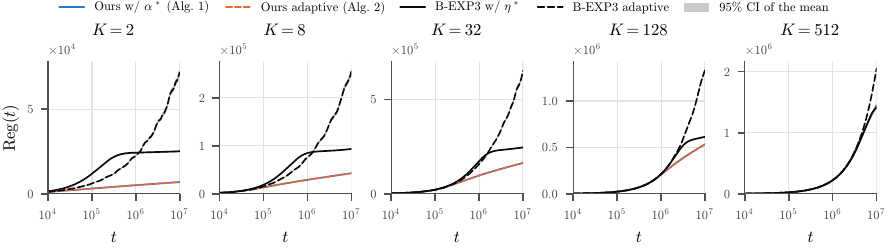}
\caption{Ablation in the number of candidate policies: Regrets across $K\in\{2,8,32,128,512\}$. Bands show 95\% CIs.}
\label{fig:e3curves}
\end{figure*}

\begin{figure}[t]
\centering
\includegraphics[width=\columnwidth]{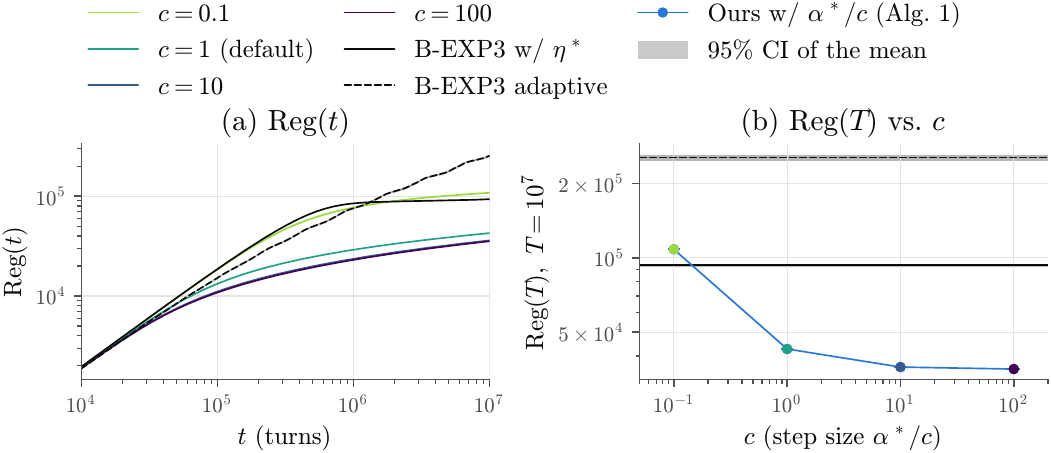}
\caption{Regret across learning rate multipliers $c$, with B-EXP3 (both tuned $\eta^*$ and adaptive) shown for reference. Bands indicate 95\% CIs.}
\label{fig:e4}
\end{figure}

We further explore through simulations: 
(i) Given enough interactions, how does our method compare asymptotically with the most naive adaptive baseline, task-level EXP3? 
(ii) How sensitive is our method to hint accuracy, the number of candidate policies, the learning rate? 

\paragraph{Setup.}
Unless stated otherwise, we evaluate $K{=}8$ policies over maximum task length $L{=}128$, with each task's length drawn from $L_b\sim\mathrm{U}\{1,\dots,128\}$. 
In task $b$, policy $k$ earns per-turn rewards $r_{t,k}=\mathrm{clip}(h_b+d_k+0.1\,\varepsilon_{t,k},0,1)$ with $\varepsilon_{t,k}\overset{\text{i.i.d.}}{\sim}\mathcal N(0,1)$, where the task difficulty $h_b\sim U[0.2,0.5]$ is shared by all policies and the offsets $d_k$ (randomly permuted per seed) give the best policy $d=\Delta=0.3$ and spread the other $K-1$ evenly over $[0,\Delta/2]$. The best policy thus leads the second-best by $\Delta/2=0.15$ and the worst by $\Delta$ in expected reward per turn.
Based on the per-turn true rewards, hints are generated as $m=(1-w)\,r+w\,u$ with $u_{t,k}\sim U[0,1]$ i.i.d., where $w{=}0$ corresponds to perfect hints and $w{=}1$ represents pure noise. We set $\beta{=}6L$. \cref{alg:aggregate-hints} uses the theoretical schedule $\alpha^{*}=\sqrt{24V}+\beta\sqrt K$ given an oracle $V$, whereas \cref{alg:aggregate-adaptive} adapts $\alpha$ dynamically via potential-gap doubling. We compare against task-level EXP3 (B-EXP3) in two variants: one where $\eta^{*}$ is tuned to the horizon $T$ using an oracle $\sum_b L_b^2$, and an anytime version utilizing the doubling trick. 
All results report the mean and 95\% confidence intervals (CIs) across 50 random seeds.

\paragraph{Asymptotic performance.}
Across a horizon of $T{=}10^8$ (Figure~\ref{fig:e1}, Table~\ref{tab:e1}), Algorithm~\ref{alg:aggregate-adaptive} consistently outperforms tuned B-EXP3 for all $t\ge10^4$ and beats the doubling  baseline from $t\approx5\times10^4$ onward, regardless of $w$. 
Our regret exhibits $\log T$ scaling, whereas the adaptive B-EXP3 with doubling trick scales as $\sqrt T$. 
By $T{=}10^8$, our Algorithm~\ref{alg:aggregate-adaptive} reaches a regret of $5.55\times 10^4$, an order of magnitude lower compared to $2.87\times 10^5$ for B-EXP3 with $\eta^*$ and $8.43\times 10^5$ for adaptive B-EXP3 using doubling trick. 
Algorithm~\ref{alg:aggregate-hints} matches Algorithm~\ref{alg:aggregate-adaptive} when $w{=}0$, but degrades as hint errors inflate the $\sqrt{24V}$ term. 
In sum, given sufficient interactions, our approach incurs only a fraction of the regret of B-EXP3 with $\eta^*$. 

\begin{table}[t]
\centering\small
\setlength{\tabcolsep}{4pt}
\caption{Terminal regret at $T{=}10^8$ with varied levels of hint accuracy. Regret values are reported as mean $\pm$ 95\% CI over 50 seeds, in units of$\times10^{3}$ due to limited space. 
}
\label{tab:e1}
\begin{tabular}{@{}lrrrr@{}}
\toprule
$w$ & Ours $\alpha^*$ & Ours adaptive & B-EXP3 w/ $\eta^*$ & B-EXP3 adaptive\\
\midrule
0   & $55.50\pm0.00$  & $55.50\pm0.00$ & \multirow{3}{*}{$287.5\pm2.4$} & \multirow{3}{*}{$843\pm22$}\\
0.1 & $148.30\pm0.04$ & $55.50\pm0.02$ & & \\
0.2 & $240.79\pm0.12$ & $55.50\pm0.04$ & & \\
\bottomrule
\end{tabular}
\end{table}

\paragraph{Ablating hint accuracy.}
We vary the noise level injected into the hints to explore how our method performs with varied hint accuracies. 
Evaluating $w\in\{0.0,0.2,0.4,0.6,0.8,1.0\}$ at terminal time $T{=}10^7$ (Figures~\ref{fig:e2curves}), Our Algorithm~\ref{alg:aggregate-adaptive} remains approximately invariant to hint noise, substantially outperforming both B-EXP3 variants by almost an order of magnitude in terminal regrets, even under pure-noise hints. 
However, Algorithm~\ref{alg:aggregate-hints}'s regret approximately increases linearly with $w$, outperforming B-EXP3 with $\eta^*$ only near $w{=}0$ and the doubling to $w{=}0.6$. 
These results suggests that our Algorithm~\ref{alg:aggregate-adaptive} is highly resilient to degraded hints, whereas the theoretically tuned Algorithm~\ref{alg:aggregate-hints} relies on accurate hints, mostly because its theoretically optimal learning rate hinges on hint accuracy.

\paragraph{Ablating number of candidate policies.}
To explore how our algorithms perform asymptotically when facing different number of candidate policies, 
we fix perfect hints ($w{=}0$) vary the policy set size across $K\in\{2,8,32,128,512\}$, measuring regrets up to $T{=}10^7$. 
The resulting regret curves are in Figure~\ref{fig:e3curves}. 
Our regret scales from $6.9\times 10^3$ ($K{=}2$) to $1.45\times 10^6$ ($K{=}512$), compared to $2.51 \times 10^4$ and $1.41 \times 10^6$ for B-EXP3 with $\eta^*$. 
Our method outperforms tuned B-EXP3 for all $K\le128$, trailing by less than $3\%$ at $K{=}512$, and beats B-EXP3 with doubling across all $K$ ($0.10$--$0.71\times$). 
The main trend here is that our relative advantage narrows as $K$ increases, matching the theoretical bound. 

\paragraph{Ablating learning rate.}
We scale the learning rate by a factor $c\in\{0.1,1,10,100\}$, i.e., setting $\alpha^{*}/c$ in \cref{alg:aggregate-hints}, with $T{=}10^7$ and $w{=}0$. 
Regret curves are in Figure~\ref{fig:e4}. 
Larger learning rates yield modest gains with diminishing returns, whereas reducing the rate ten-fold degrades performance by $2.5\times$, pushing regret above B-EXP3 with $\eta^*$. 
These show that our performance is robust across three orders of magnitude, and the theoretical $\alpha^{*}$ is conservative when hints are accurate.

\paragraph{Summary of findings.}
Given sufficient interactions, our method achieves $\log T$ rather than $\sqrt T$ regret growth, outperforming even the tuned B-EXP3 method. 
Algorithm~\ref{alg:aggregate-adaptive} appears to be less sensitive to hint noise and more robust to learning-rate scaling while its advantage narrows as the number of candidate policies grows.  

\section{Conclusion}
We introduced an online learning framework that dynamically selects activation steering policies for multi-turn LLM agents by pairing intermediate behavioral hints with delayed task-level aggregate feedback. 
By bridging the gap between turn-level decisions and task-level aggregate rewards, our approach offers provable learning guarantees and consistently outperforms static and heuristic steering baselines on realistic coding benchmarks. 
We believe this work provides a practical foundation for guiding autonomous LLM agents through complex, multi-step software and Web development workflows.

\clearpage

\bibliographystyle{plainnat}
\bibliography{WWW2027/refs}

\appendix

\section{Auxiliary Lemmas for Theorem~\ref{thm:aggregate-fixed}}
\label{app:optimistic-proofs}
\label{app:fixed-steering-proof}
This appendix establishes the auxiliary lemmas used in the proof of Theorem~\ref{thm:aggregate-fixed} in Section~\ref{sec:hint-method}. Throughout, the reward vectors, hint vectors, and task partition are fixed independently of all learner randomization.

We first compute the action marginals and the probability of choosing one policy throughout a task. The formula for selecting the same policy throughout a task generalizes Proposition~1 of Lorberbom et al.~\cite{lorberbom2021gumbel} from two distributions to all turns of a task.
\begin{lemma}
\label{lem:steering-coupling}
For task $b$, let $p_t\in\Delta_K$, $t\in\mathcal I_b$, be $\mathcal F_{b-1}$-measurable with $p_{t,k}>0$ for all $t\in\mathcal I_b$ and $k\in[K]$. Write $\xi_k=\xi_{b,k}$ for independent $\operatorname{Exp}(1)$ draws, independent of $\mathcal F_{b-1}$. At each turn, set $A_t=\allowbreak\arg\min_{k\in[K]}\xi_k/p_{t,k}$. Then $\mathbb P_b(A_t=k)=p_{t,k}$ for every $t\in\mathcal I_b$ and $k\in[K]$, and
\begin{equation}\label{eq:app-coincidence}
\mathbb P_b(A_{s_b}=\cdots=A_{e_b}=k)=\rho_{b,k}
=\left(\sum_{\ell=1}^K\max_{t\in\mathcal I_b}
\frac{p_{t,\ell}}{p_{t,k}}\right)^{-1}.
\end{equation}
\end{lemma}
\begin{proof}
Condition throughout on $\mathcal F_{b-1}$, so the $p_t$ are fixed and the $\xi_k$ remain independent $\operatorname{Exp}(1)$ variables. For a fixed turn $t$, the variables $\xi_k/p_{t,k}$ are independent exponentials with rates $p_{t,k}$. Since $\sum_kp_{t,k}=1$, their minimum has index $k$ with probability $p_{t,k}$, proving the marginal identity. Ties have probability zero.

For the joint event, condition further on $\xi_k=z$. Up to ties, all actions equal $k$ exactly when
$\xi_\ell\geq\allowbreak z\max_{t\in\mathcal I_b}(p_{t,\ell}/p_{t,k})$ for every $\ell\ne k$. Using the definition of $\rho_{b,k}$, independence and integration give
$\mathbb P_b(A_{s_b}=\cdots=A_{e_b}=k)\allowbreak=\int_0^\infty e^{-z/\rho_{b,k}}\,dz\allowbreak=\rho_{b,k}$.
\end{proof}

We next establish the relative stability of the within-task distributions.

\begin{lemma}\label{lem:steering-stability}
For task $b$, let $q=Q_{\alpha,\beta}(G_{b-1})$ and $\beta=6L$. On the hint segments $G_{b-1}+v$, where $v=\sum_{i=s_b}^{t-1}m_i+s m_t$, $t\in\mathcal I_b$, and $0\leq s\leq t-s_b+1$, we have
\begin{equation}\label{eq:app-relative}
\frac45q_k\leq Q_{\alpha,\beta}(G_{b-1}+v)_k\leq\frac65q_k.
\end{equation}
On the residual segments, for $0\leq s\leq1$,
\begin{equation}\label{eq:app-relative-residual}
\frac12q_k\leq Q_{\alpha,\beta}(G_{b-1}+M_b+s\widehat D_b)_k\leq\frac32q_k.
\end{equation}
Moreover, the probabilities in Algorithm~\ref{alg:aggregate-hints} satisfy
\begin{equation}\label{eq:app-coincidence-lower}
\rho_{b,k}\geq\frac23q_k.
\end{equation}
\end{lemma}
\begin{proof}
We first establish a general perturbation bound. For $p=Q_{\alpha,\beta}(G_{b-1}+v)$ and $\varepsilon=\|q\odot v\|_2/\beta<1$, we claim that $\left\|(p-q)\oslash q\right\|_2\allowbreak\leq\varepsilon/(1-\varepsilon)$, where $\odot$ and $\oslash$ denote coordinatewise multiplication and division, respectively. The barrier diverges at the simplex boundary, and $\psi_{\alpha,\beta}$ is strictly convex on its relative interior. Hence $Q_{\alpha,\beta}$ exists uniquely and has positive coordinates. The KKT equations for $p$ and $q$, subtracted and paired with $p-q$, give
\[
\langle v,p-q\rangle
=\langle\nabla\psi_{\alpha,\beta}(p)-\nabla\psi_{\alpha,\beta}(q),p-q\rangle.
\]
The Lagrange multiplier term vanishes because $\sum_k(p_k-q_k)=0$. The Tsallis contribution is nonnegative by convexity. With $u_k=(p_k-q_k)/q_k$, the barrier contribution is $\beta\sum_k u_k^2/(1+u_k)$. Therefore
\[
\begin{aligned}
\frac{\beta\|u\|_2^2}{1+\|u\|_2}
&\leq\beta\sum_k\frac{u_k^2}{1+u_k}
\leq\langle q\odot v,u\rangle
\leq\beta\varepsilon\|u\|_2.
\end{aligned}
\]
If $u=0$ the assertion is immediate; otherwise division and rearrangement give the stated inequality.

The simplex constraint gives shift invariance: $Q_{\alpha,\beta}(x+c\mathbf1)=Q_{\alpha,\beta}(x)$ for any $c\in\mathbb R$. We may therefore apply the perturbation bound to $v-c\mathbf1$. On each hint segment, the coordinates of $v$ lie in $[0,2L_b]$. Centering by $L_b\mathbf1$ gives $\|q\odot(v-L_b\mathbf1)\|_2\leq L_b$, so $\varepsilon\leq1/6$ and the relative deviation is at most $1/5$. This proves~\eqref{eq:app-relative}, including its application to every $p_t$. Lemma~\ref{lem:steering-coupling} then gives
\[
\rho_{b,k}^{-1}
=\sum_{\ell=1}^K\max_{t\in\mathcal I_b}\frac{p_{t,\ell}}{p_{t,k}}
\leq\frac{3}{2q_k}\sum_{\ell=1}^Kq_\ell
=\frac{3}{2q_k},
\]
which proves~\eqref{eq:app-coincidence-lower}.

If the actions change within the task, $\widehat D_b=0$. Otherwise, all actions equal some $k$, and $|Z_b|\leq L_b$ gives
$
\|q\odot\widehat D_b\|_2
=\frac{q_k|Z_b|}{\rho_{b,k}}
\leq\frac32L_b.
$
Since every coordinate of $M_b$ lies in $[0,L_b]$, centering the residual perturbation gives, for $0\leq s\leq1$,
\[
\left\|q\odot\left(M_b+s\widehat D_b-\frac{L_b}{2}\mathbf1\right)\right\|_2
\leq\frac{L_b}{2}+\frac{3L_b}{2}=2L_b.
\]
Thus $\varepsilon\leq1/3$ and the relative deviation is at most $1/2$, proving~\eqref{eq:app-relative-residual}. These shifts are used only in the analysis and do not change the algorithm.
\end{proof}

The next lemma gives the curvature bounds used in the hint and residual analyses. For a positive-semidefinite matrix $W$, write $\|z\|_W^2=z^\top Wz$. The matrix $W(q)$ below provides a common seminorm for all update segments within the task.
\begin{lemma}\label{lem:steering-hessian}
For $x\in\mathbb R^K$ and $p=Q_{\alpha,\beta}(x)$, define
\[
 v_k(p)=\frac{p_k^2}{\beta+(\alpha/2)\sqrt{p_k}},\quad
 W(p)=\operatorname{diag}(v(p))
 -\frac{v(p)v(p)^\top}{\sum_kv_k(p)}.
\]
Then $\nabla\Phi_{\alpha,\beta}(x)=p$, $\nabla^2\Phi_{\alpha,\beta}(x)=W(p)$, and
\begin{equation}\label{eq:app-centered}
\|z\|_{W(p)}^2=\min_{c\in\mathbb R}\sum_kv_k(p)(z_k-c)^2.
\end{equation}
Writing $W=W(q)$, the Hessians on the hint segments satisfy $\tfrac12W\preceq\nabla^2\Phi_{\alpha,\beta}\preceq2W$, while those on the residual segments satisfy $\nabla^2\Phi_{\alpha,\beta}\preceq\tfrac94W$.
\end{lemma}
\begin{proof}
The regularizer has diagonal Hessian with entries $\beta/p_k^2+\alpha/(2p_k^{3/2})$. Differentiating its KKT equations, subject to $\sum_k dp_k=0$, gives $dp_k=v_k(p)(dx_k-d\lambda)$, where $d\lambda=\frac{\sum_kv_k(p)dx_k}{\sum_kv_k(p)}$.
This yields the Hessian formula; differentiating the optimized value yields $\nabla\Phi_{\alpha,\beta}=Q_{\alpha,\beta}$. The right side of~\eqref{eq:app-centered} is minimized at the weighted mean of the coordinates of $z$, which gives the identity.

It remains to compare the Hessians along the update segments. For a fixed coordinate $k$, set $a=p_k/q_k$. Direct substitution gives $\frac{v_k(p)}{v_k(q)}\allowbreak=a^2\frac{\beta+(\alpha/2)\sqrt{q_k}}{\beta+(\alpha/2)\sqrt{a}\sqrt{q_k}}$.
For $a\leq1$, the ratio lies in $[a^2,a^{3/2}]$; for $a\geq1$, these bounds are reversed. Lemma~\ref{lem:steering-stability} places it in $[16/25,36/25]$, and hence in $[1/2,2]$, on the hint segments; on the residual segments it is at most $9/4$. Applying these coordinatewise bounds before minimizing over $c$ proves the stated matrix comparisons. In particular, $W$ annihilates constant vectors; all subsequent estimates use this centered seminorm.
\end{proof}

The following lemma shows that $\widehat D_{b,k}$ is an unbiased estimator of $D_{b,k}$ and bounds its second moment.

\begin{lemma}\label{lem:steering-moments}
Let $V_b=L_b\sum_{t\in\mathcal I_b}\|r_t-m_t\|_\infty^2$. For each policy $k\in[K]$, the estimator satisfies
\begin{equation}\label{eq:app-second-moment}
\mathbb E_b[\widehat D_{b,k}]=D_{b,k},\,
\mathbb E_b[\widehat D_{b,k}^{\,2}]\leq\frac{3D_{b,k}^2}{2q_k},\,
\mathbb E_b\!\left[\sum_kv_k(q)\widehat D_{b,k}^{\,2}\right]
\leq\frac{4\sqrt K}{\alpha}V_b.
\end{equation}
\end{lemma}
\begin{proof}
Fix $k\in[K]$. By Lemma~\ref{lem:steering-coupling}, all actions in task $b$ equal $k$ with conditional probability $\rho_{b,k}$. The estimator is $D_{b,k}/\rho_{b,k}$ on this event and zero otherwise. Hence $\mathbb E_b[\widehat D_{b,k}]=D_{b,k}$ and $\mathbb E_b[\widehat D_{b,k}^{\,2}]\allowbreak=\frac{D_{b,k}^2}{\rho_{b,k}}\allowbreak\leq\frac{3D_{b,k}^2}{2q_k}$, where the inequality follows from Lemma~\ref{lem:steering-stability}.

For~\eqref{eq:app-second-moment}, Cauchy--Schwarz gives $D_{b,k}^2\allowbreak\leq L_b\sum_{t\in\mathcal I_b}(r_{t,k}-m_{t,k})^2\allowbreak\leq V_b$. Using $v_k(q)\leq2q_k^{3/2}/\alpha$ from Lemma~\ref{lem:steering-hessian} and summing over $k$, we obtain
\[
\begin{aligned}
\mathbb E_b\!\left[\sum_kv_k(q)\widehat D_{b,k}^{\,2}\right]
\leq\frac{3V_b}{\alpha}\sum_k\sqrt{q_k}\leq\frac{3\sqrt K}{\alpha}V_b
\leq\frac{4\sqrt K}{\alpha}V_b.
\end{aligned}
\]
Here $\sum_k\sqrt{q_k}\leq\sqrt K$ by Cauchy--Schwarz and $\sum_kq_k=1$.
\end{proof}

We next show how the gain from the scaled hint controls the discrepancy between the turn-specific and task-final distributions.
\begin{lemma}\label{lem:steering-surplus}
Fix a task $b$. For $t\in\mathcal I_b$, define $d_t=r_t-m_t$, $x_t=G_{b-1}+\sum_{s=s_b}^t m_s$, and $a_t=Q_{\alpha,\beta}(x_t)$, with $x_{s_b-1}=G_{b-1}$ and $a_{s_b-1}=q$. Let $\sigma_t=\langle p_t,m_t\rangle-\Phi_{\alpha,\beta}(x_t)+\Phi_{\alpha,\beta}(x_{t-1})$. Then
\begin{equation}\label{eq:app-surplus}
\sigma_t\geq\frac{t-s_b+1}{4}\|m_t\|_W^2\geq0,
\end{equation}
and
\begin{equation}\label{eq:app-cancellation}
\sum_{t\in\mathcal I_b}\langle a_{e_b}-p_t,d_t\rangle-\sum_{t\in\mathcal I_b}\sigma_t
\leq\frac{32}{\alpha}V_b.
\end{equation}
\end{lemma}
\begin{proof}
Here $x_{e_b}=G_{b-1}+M_b$ and $p_t=Q_{\alpha,\beta}(x_{t-1}+(t-s_b+1)m_t)$. We first prove~\eqref{eq:app-surplus}. Applying the fundamental theorem of calculus twice and the lower Hessian bound in Lemma~\ref{lem:steering-hessian} gives
\[
\begin{aligned}
\sigma_t
=\int_0^1\!\int_u^{t-s_b+1}
m_t^\top\nabla^2\Phi_{\alpha,\beta}(x_{t-1}+v m_t)m_t\,dv\,du \geq \frac12\left(t-s_b+\frac12\right)\|m_t\|_W^2
\geq\frac{t-s_b+1}{4}\|m_t\|_W^2.
\end{aligned}
\]
To prove~\eqref{eq:app-cancellation}, define $\widetilde D_{s_b-1}=0$ and $\widetilde D_t=\sum_{s=s_b}^t d_s$ within this proof, so $\widetilde D_{e_b}=D_b$. Summation by parts gives
\[
\sum_{t\in\mathcal I_b}\langle a_{e_b}-p_t,d_t\rangle=\sum_{t\in\mathcal I_b}\langle a_t-a_{t-1},\widetilde D_t\rangle
-\sum_{t\in\mathcal I_b}\langle p_t-a_{t-1},d_t\rangle.
\]
Indeed, expanding $\widetilde D_t=\widetilde D_{t-1}+d_t$ on the right makes its $a_t$ terms telescope to $\langle a_{e_b},\widetilde D_{e_b}\rangle$. The two differences are integrals of local Hessians, over segments with parameter ranges $[0,1]$ and $[0,t-s_b+1]$, respectively. Positive-semidefinite Cauchy--Schwarz and Lemma~\ref{lem:steering-hessian} bound their absolute inner products by $2\|m_t\|_W\|\widetilde D_t\|_W$ and $2(t-s_b+1)\|m_t\|_W\|d_t\|_W$. Combining these bounds with~\eqref{eq:app-surplus} and applying Young's inequality gives
\begin{align*}
\sum_{t\in\mathcal I_b}\langle a_{e_b}-p_t,d_t\rangle-\sum_{t\in\mathcal I_b}\sigma_t\leq8\sum_{t\in\mathcal I_b}\frac{\|\widetilde D_t\|_W^2}{t-s_b+1}
+8\sum_{t\in\mathcal I_b}(t-s_b+1)\|d_t\|_W^2 \leq16L_b\sum_{t\in\mathcal I_b}\|d_t\|_W^2.
\end{align*}
The last step uses $\|\widetilde D_t\|_W^2\leq(t-s_b+1)\sum_{s=s_b}^t\|d_s\|_W^2$ and $t-s_b+1\leq L_b$. Finally, by choosing $c=0$ in~\eqref{eq:app-centered},
\[
\|d_t\|_W^2
\leq\sum_kv_k(q)d_{t,k}^2
\leq\frac2\alpha\|d_t\|_\infty^2\sum_kq_k^{3/2}
\leq\frac2\alpha\|d_t\|_\infty^2.
\]
This proves~\eqref{eq:app-cancellation}.
\end{proof}

The following lemma controls the potential gap. 

\begin{lemma}\label{lem:steering-gap}
For task $b$, let $\delta_b$ be the potential gap defined in~\eqref{eq:aggregate-potential-gap}, evaluated with parameters $(\alpha,\beta)$. Then
\begin{equation}\label{eq:app-gap-bounds}
\mathbb E_b[\delta_b]\leq\frac{48\sqrt K}{\alpha}V_b,\quad \operatorname{Var}(\delta_b\mid\mathcal F_{b-1})\leq\frac{25}{16}V_b, \quad \delta_b\leq\frac54L_b.
\end{equation}
\end{lemma}
\begin{proof}
Fix task $b$ and use $d_t$, $x_t$, $a_t$, and $\sigma_t$ from Lemma~\ref{lem:steering-surplus}. We first bound the conditional mean. Taylor expansion along the residual segment yields
\[
\Phi_{\alpha,\beta}(x_{e_b}+\widehat D_b)
=\Phi_{\alpha,\beta}(x_{e_b})+\langle a_{e_b},\widehat D_b\rangle+\mathcal R_b,
\]
where
\[
\begin{aligned}
\mathcal R_b
=\int_0^1(1-u)\widehat D_b^\top
\nabla^2\Phi_{\alpha,\beta}(x_{e_b}+u\widehat D_b)\widehat D_b\,du\leq\frac98\|\widehat D_b\|_W^2
\leq\frac98\sum_kv_k(q)\widehat D_{b,k}^{\,2}.
\end{aligned}
\]
Lemma~\ref{lem:steering-hessian} bounds the remainder. The marginals in Lemma~\ref{lem:steering-coupling} give $\mathbb E_b[Z_b]=\allowbreak\sum_{t\in\mathcal I_b}\langle p_t,d_t\rangle$. Substituting the expansion into the definition of $\delta_b$, taking conditional expectations, and applying Lemmas~\ref{lem:steering-surplus} and~\ref{lem:steering-moments} gives
\[
\begin{aligned}
\mathbb E_b[\delta_b]
&=-\sum_{t\in\mathcal I_b}\sigma_t+\sum_{t\in\mathcal I_b}\langle a_{e_b}-p_t,d_t\rangle
+\mathbb E_b[\mathcal R_b]\leq\frac{32+(9/2)\sqrt K}{\alpha}V_b
\leq\frac{48\sqrt K}{\alpha}V_b.
\end{aligned}
\]
For the variance and maximum increase, separate the deterministic hint contribution by defining
\[
c_b=\Phi_{\alpha,\beta}(x_{e_b})-\Phi_{\alpha,\beta}(G_{b-1})-\sum_{t\in\mathcal I_b}\langle p_t,m_t\rangle
=-\sum_{t\in\mathcal I_b}\sigma_t\leq0.
\]
Conditional on $\mathcal F_{b-1}$, $c_b$ is deterministic. If the actions change within the task, $\widehat D_b=0$ and $\delta_b-c_b=-Z_b$. If every action equals $k$, integration of $\nabla\Phi_{\alpha,\beta}=Q_{\alpha,\beta}$ gives
\[
\delta_b-c_b
=Z_b\left(\int_0^1
\frac{\bigl[Q_{\alpha,\beta}(x_{e_b}+u\widehat D_b)\bigr]_k}{\rho_{b,k}}\,du-1\right).
\]
By Lemma~\ref{lem:steering-stability}, $0<\frac{\bigl[Q_{\alpha,\beta}(x_{e_b}+u\widehat D_b)\bigr]_k}{\rho_{b,k}}\allowbreak\leq\frac32\cdot\frac32=\frac94$.
The factor in parentheses has absolute value at most $5/4$. Thus, on every trajectory,
$
|\delta_b-c_b|\leq\frac54|Z_b|
\leq\frac54\min\{L_b,\sqrt{V_b}\},
$
where the last inequality follows from $\|d_t\|_\infty\leq1$ and Cauchy--Schwarz. Since $c_b\leq0$ is $\mathcal F_{b-1}$-measurable, we obtain $\delta_b\leq5L_b/4$ and $\operatorname{Var}(\delta_b\mid\mathcal F_{b-1})\allowbreak\leq\mathbb E_b[(\delta_b-c_b)^2]\allowbreak\leq25V_b/16$.
\end{proof}

\section{Proof of Theorem~\ref{thm:aggregate-adaptive}}
\label{app:adaptive-steering-proof}
\begin{proof}
Fix $k\in[K]$ and use $u=(1-\gamma)e_k+\gamma\mathbf1/K$, with $\gamma=\beta K/(T+\beta K)$, throughout the run. Recall $V$ from Theorem~\ref{thm:aggregate-fixed}. Let $J$ be the number of epochs, including a final empty epoch if the last task triggers a restart. Their thresholds are $H^{(i)}=2^iH_0$, $i=0,\ldots,J-1$, where $H_0=4\beta K$, and $H_{\mathrm f}=H_B=H^{(J-1)}$. Thus $J=1+\log_2(H_{\mathrm f}/H_0)$.

Each epoch starts from score zero and uses $\alpha^{(i)}=H^{(i)}/(4\sqrt K)$. By Lemma~\ref{lem:steering-gap}, a completed epoch's cumulative gap is at most $H^{(i)}+5L/4\leq2H^{(i)}$; the final epoch's gap is at most $H_{\mathrm f}$. Together with~\eqref{eq:app-comparator-cost} and $\sum_iH^{(i)}<2H_{\mathrm f}$, this gives
\[
\begin{aligned}
\sum_i\psi_{\alpha^{(i)},\beta}(u)
&\leq H_{\mathrm f}+J\beta K\log(1+T/(\beta K)),\\
\sum_{b=1}^B\delta_b
&\leq2\sum_{i=0}^{J-2}H^{(i)}+H_{\mathrm f}\leq3H_{\mathrm f}.
\end{aligned}
\]
Apply~\eqref{eq:app-potential-regret} in each realized epoch and sum before taking expectations. Since $G_{b-1}$ and $\alpha_b$ are $\mathcal F_{b-1}$-measurable, Lemmas~\ref{lem:steering-coupling} and~\ref{lem:steering-moments} apply to every task. Using the same $u$ across epochs and adding its smoothing cost yields
\begin{equation}\label{eq:app-epoch-accounting}
\operatorname{Reg}_T\leq4\mathbb E[H_{\mathrm f}]+\beta K\log(1+T/(\beta K))(\mathbb E[J]+1).
\end{equation}

To bound $H_{\mathrm f}$, define the martingale $N_j=\sum_{b=1}^j(\delta_b-\mathbb E_b[\delta_b])$, $j=0,\ldots,B$, and let $N^*=\max_{0\leq j\leq B}|N_j|$. Lemma~\ref{lem:steering-gap} and orthogonality of martingale increments give $\mathbb E[N_B^2]\leq25V/16$. Doob's $L^2$ maximal inequality and Cauchy--Schwarz then yield
\begin{equation}\label{eq:app-martingale-max}
\mathbb E[N^*]\leq2\sqrt{\mathbb E[N_B^2]}\leq\frac52\sqrt V.
\end{equation}
If no restart occurs, $H_{\mathrm f}=H_0$. Otherwise, the last completed epoch has threshold $H=H_{\mathrm f}/2$, coefficient $\alpha=H/(4\sqrt K)$, and tasks $a,\ldots,d$. The restart rule and Lemma~\ref{lem:steering-gap} give
\[
H<\sum_{b=a}^d\delta_b
=\sum_{b=a}^d\mathbb E_b[\delta_b]+N_d-N_{a-1}
\leq\frac{192KV}{H}+2N^*.
\]
This holds for every realized interval, including the random endpoints $a,d$. Solving the quadratic inequality and including the no-restart case gives
\begin{equation}\label{eq:app-final-threshold}
H_{\mathrm f}\leq4\beta K+4N^*+2\sqrt{192KV}.
\end{equation}
Consequently, $\mathbb E[H_{\mathrm f}]=\mathcal{O}(\beta K+\sqrt{KV})$, and Jensen's inequality yields
\[
\mathbb E[J]\leq1+\log_2\!\left(\frac{\mathbb E[H_{\mathrm f}]}{4\beta K}\right)
=\mathcal{O}\!\left(1+\log\!\left(1+\frac{\sqrt{KV}}{\beta K}\right)\right).
\]
Substitute these estimates into~\eqref{eq:app-epoch-accounting}. For $T\geq\beta K$, absorb the additive $\beta K$ into the logarithmic term; for $T<\beta K$, use $\operatorname{Reg}_T\leq T\leq2\beta K\log(1+T/(\beta K))$. With $\beta=6L$, this gives
\begin{equation}\label{eq:app-adaptive-sharp}
\operatorname{Reg}_T=\mathcal{O}\Bigg(\sqrt{KV}+KL\log\!\left(1+\frac{T}{KL}\right)\left[1+\log\!\left(1+\frac{\sqrt{KV}}{KL}\right)\right]\Bigg).
\end{equation}
Finally, $V\leq LT$ and $1+\log(1+\sqrt{x})=\mathcal{O}(1+\log(1+x))$ for $x\geq0$ give~\eqref{eq:aggregate-adaptive-bound}.
\end{proof}

\section{Proof of Theorem~\ref{thm:steering-lower-bound}}
\label{app:steering-lower-bound}
\begin{proof}[Proof of Theorem~\ref{thm:steering-lower-bound}]
Set $\varepsilon=\sqrt{\overline V/(BL^2)}\in(0,1/2]$ and $\gamma=\tfrac12\sqrt{K/B}\leq1/2$. We construct a distribution over fixed instances satisfying $V\leq\overline V$ and show that the learner's expected regret, averaged over this distribution, is at least $\sqrt{K\overline V}/32$.

Choose $h$ uniformly from $[K]$ and draw $U_1,\ldots,U_B$ independently from $\operatorname{Unif}[0,1]$ before interaction begins, independently of the learner's randomization. For every $t\in\mathcal I_b$ and $k\in[K]$, let
\begin{equation}\label{eq:lower-hard-instance}
m_{t,k}=\tfrac12, \quad r_{t,k}=\tfrac12+\varepsilon\bigl(2U_b-1
             +\gamma U_b(1-U_b)\mathbb{I}\{k=h\}\bigr).
\end{equation}
The function $z\mapsto2z-1+\gamma z(1-z)$ is increasing on $[0,1]$, since its derivative is $2+\gamma(1-2z)>0$, and its endpoint values are $-1$ and $1$. Thus all rewards lie in $[0,1]$ and $\|r_t-m_t\|_\infty\leq\varepsilon$. Every realization therefore satisfies $V\leq BL^2\varepsilon^2=\overline V$. Policy $h$ is a best policy at every turn, hence also a best fixed policy in hindsight. All hints are identical and reveal no information about $h$.

Let $x_{b,k}$ be the fraction of task $b$ assigned to policy $k$:
$
x_{b,k}=\frac1L\sum_{t\in\mathcal I_b}\mathbb{I}\{A_t=k\}.
$ Thus $x_b\in\Delta_K$; no restriction to a single policy within a task is imposed. Conditional on the past history and the learner's random seed, the entire action sequence in task $b$ is determined before its aggregate feedback: the hints are constant and no rewards are observed within the task. In particular, that sequence is independent of the fresh $U_b$.

To compare the feedback distributions, apply the following invertible normalization to the observed aggregate reward:
\begin{equation}\label{eq:lower-normalized-observation}
Y_b=\frac{R_b-L(1/2-\varepsilon)}{2\varepsilon L}
    =U_b+\frac\gamma2 x_{b,h}U_b(1-U_b).
\end{equation}
Consider also a null experiment with the same hints and rewards $r_{t,k}=1/2+\varepsilon(2U_b-1)$ for all policies. In that experiment $Y_b=U_b$.

To quantify the difference between these observation laws, write $f_c(z)=z+cz(1-z)$ for $0\leq c\leq1/4$. This is a continuously differentiable bijection of $[0,1]$ onto itself, with $f_c'(z)=1+c(1-2z)>0$. If $U$ is uniform, the density of $f_c(U)$ at $y$ is $1/f_c'(f_c^{-1}(y))$. A change of variables therefore gives
\begin{equation}\label{eq:lower-one-task-kl}
\begin{aligned}
&\operatorname{KL}\!\left(\operatorname{Unif}[0,1]\,
                 \middle\|\,\mathcal L(f_c(U))\right)=\int_0^1 f_c'(z)\log f_c'(z)\,dz
\leq\frac{c^2}{6(1-c)}\leq\frac{c^2}{3}.
\end{aligned}
\end{equation}
Here $\mathcal L$ denotes the probability law and $\operatorname{KL}$ denotes relative entropy. For the first inequality, put $v=c(1-2z)$ and use
$
(1+v)\log(1+v)-v\leq\frac{v^2}{2(1-c)}, |v|\leq c.
$ This follows by Taylor's theorem because the second derivative of $(1+v)\log(1+v)-v$ is $1/(1+v)\leq1/(1-c)$. Integrating uses $\int_0^1c(1-2z)\,dz=0$ and $\int_0^1c^2(1-2z)^2\,dz=c^2/3$.

Let $P_h$ and $P_0$ be the laws of the learner's random seed and full observation history under hidden policy $h$ and the null experiment, respectively, and let $\mathbb E_h$ and $\mathbb E_0$ denote expectations under these laws. The seed has the same distribution in both experiments. Given the seed and past observations, the action sequence is the same deterministic function in both experiments, so only the conditional laws of $Y_b$ contribute to the relative-entropy chain rule. Applying~\eqref{eq:lower-one-task-kl} with $c=\gamma x_{b,h}/2\leq1/4$ yields
$\operatorname{KL}(P_0\|P_h)\allowbreak\leq\frac{\gamma^2}{12}\mathbb E_0[\sum_{b=1}^B x_{b,h}^2]$.
Since $\sum_{h=1}^Kx_{b,h}^2\leq1$ on every trajectory,
\begin{equation}\label{eq:lower-average-kl}
\frac1K\sum_{h=1}^K\operatorname{KL}(P_0\|P_h)
\leq\frac{\gamma^2B}{12K}.
\end{equation}
Define $X_h=\sum_{b=1}^B x_{b,h}\in[0,B]$. Pinsker's inequality, Cauchy--Schwarz, and~\eqref{eq:lower-average-kl} give
\begin{equation}\label{eq:lower-best-arm-allocation}
\begin{aligned}
\frac1K\sum_{h=1}^K\mathbb E_h[X_h]
\leq\frac1K\sum_{h=1}^K\mathbb E_0[X_h]
 +\frac BK\sum_{h=1}^K\sqrt{\frac{\operatorname{KL}(P_0\|P_h)}2}\leq\frac BK+B\gamma\sqrt{\frac{B}{24K}}.
\end{aligned}
\end{equation}
The first term is $B/K$ because $\sum_hX_h=B$ on every trajectory.

In task $b$, the realized reward difference between policy $h$ and the learner is $\varepsilon\gamma L U_b(1-U_b)(1-x_{b,h})$. Conditional on $h$, the current $U_b$ is independent of the pre-feedback allocation $x_b$, and $\mathbb E[U_b(1-U_b)]=1/6$. Hence, writing $\mathcal R_h$ for the regret averaged over $U_1,\ldots,U_B$ and the learner's randomization conditional on $h$, we obtain $\mathcal R_h\allowbreak=\frac{\varepsilon\gamma L}{6}\bigl(B-\mathbb E_h[X_h]\bigr)$.
This is regret to the best fixed policy, since $h$ is pointwise best. Using~\eqref{eq:lower-best-arm-allocation}, we obtain
\[
\frac1K\sum_{h=1}^K\mathcal R_h
\geq\frac{\varepsilon\gamma LB}{6}
       \left(1-\frac1K-\gamma\sqrt{\frac{B}{24K}}\right).
\]
Since $\gamma\sqrt{B/(24K)}=1/(2\sqrt{24})\leq1/8$ and $K\geq2$, the parenthesis is at least $3/8$. Therefore
\[
\frac1K\sum_{h=1}^K\mathcal R_h
\geq\frac{\varepsilon\gamma LB}{16}
=\frac{\varepsilon L\sqrt{KB}}{32}
=\frac{\sqrt{K\overline V}}{32}.
\]
Every realization satisfies $V\leq\overline V$, so averaging yields a fixed instance with $\operatorname{Reg}_T\geq\sqrt{K\overline V}/32$, where the expectation is only over the learner's randomization. This proves~\eqref{eq:steering-budget-lower}. The argument remains valid when $B$, $L$, and $\overline V$ are known, since they are identical across all hidden-policy alternatives and the null experiment.
\end{proof}

\end{document}